\documentclass[letterpaper, 10pt, conference]{ieeeconf}
\IEEEoverridecommandlockouts
\def\DRAFT{comment this line to remove comments}
\usepackage{bm}
\usepackage{mathtools}
\usepackage{amsmath}
\usepackage{multirow}
\usepackage{tabularx}
\usepackage{balance}
\let\labelindent\relax
\usepackage{enumitem}
\usepackage[english]{babel}
\usepackage[autostyle,english=american]{csquotes}
\usepackage{graphics} 

\usepackage{tikz}
\usetikzlibrary{automata,arrows,calc,positioning}
\usepackage{hyperref}

\ifx\DRAFT\fi

\newcommand{\cmt}[4]{\ifx\DRAFT\undefined\else\colorbox{#3}{\textcolor{#4}{\small{\textsf{[\textbf{#1}: #2]}}}}\fi}

\usepackage{graphicx}
\usepackage[ruled,vlined]{algorithm2e}

\def\widthnarrow{241.14749pt}

\title{\LARGE \bf
Accurately Tracking Relative Positions of Moving Trackers
\\ based on UWB Ranging and Inertial Sensing without Anchors
}

\author{Rayan Armani and Christian Holz
\thanks{Both authors are with the Department of Computer Science at ETH Zürich, 8092 Zürich, Switzerland.
        Contact addresses are {\tt\small firstname.lastname@inf.ethz.ch}.
        }%
}

\begin{document}

\maketitle
\thispagestyle{empty}
\pagestyle{empty}

\begin{abstract}

We present a tracking system for relative positioning that can operate on entirely moving tracking nodes without the need for stationary anchors.
Each node embeds a 9-DOF magnetic and inertial measurement unit and a \emph{single-antenna} ultra-wideband radio. 
We introduce a multi-stage filtering pipeline through which our system estimates the relative layout of all tracking nodes within the group.
The key novelty of our method is the integration of a custom Extended Kalman filter (EKF) with a refinement step via multidimensional scaling (MDS).
Our method integrates the MDS output back into the EKF, thereby creating a dynamic feedback loop for more robust estimates.
We complement our method with UWB ranging protocol that we designed to allow tracking nodes to opportunistically join and leave the group.

In our evaluation with \emph{constantly moving nodes}, our system estimated relative positions with an error of 10.2\,cm (in 2D) and 21.7\,cm (in 3D), including obstacles that occluded the line of sight between tracking nodes.
Our approach requires no external infrastructure, making it particularly suitable for operation in environments where stationary setups are impractical.

\end{abstract}

\section{INTRODUCTION}

Tracking relative positions among groups of moving nodes is beneficial for several robotic and automation scenarios.
From autonomous vehicle fleets that navigate urban environments~\cite{hao2013collaborativelocalization} to groups of moving drones that need to coordinate their maneuvers~\cite{shrit2017droneswarm}, understanding the relative positions of these nodes is paramount.
Collaborative localization is also needed in dynamic settings where the environment or the relative layout of nodes is constantly changing, and where tracking infrastructure cannot be placed in the environment.

Several challenges exist in relative node tracking.
First, while positioning needs to be accurate, a deployed system must robustly allow nodes to spontaneously join or leave~\cite{cao2010decentralized}.
Second, the lack of synchronization and coordination in lieu of tracking infrastructure hampers real-time positioning~\cite{cano2023collaborativelocalization}, especially in environments that cannot accommodate traditional tracking systems (e.g., GPS-denied environments~\cite{hao2013collaborativelocalization}).
Finally, many applications require trackers to be small and light to ease attachment to objects of interest, such as flying drones, body parts, or hand-held tools.

Prior work has researched suitable tracking approaches, especially inside-out tracking systems that support moving trackers as well as point-to-point ranging techniques to track groups of nodes.
Most accurate is camera-based inside-out tracking that, in conjunction with inertial sensors (IMUs), obtainstracks poses through visual-inertial odometry~\cite{nister2004visual, weinstein2018visual}.
Less computationally expensive are IMU-only systems~\cite{ahmad2013imu,xsens}, which more recently leverage learning-based approaches to constrain the drift they accumulate over time~\cite{huang2018imu,liu2020TLIO}.

Much research has investigated wireless signals for ranging to support tracking, specifically through the use of narrow-band signals such as WiFi~\cite{wifi_2,wifi_3} and Bluetooth low energy (BLE~\cite{bluetooth_2,tracko}).
Electromagnetic-field (EM) sensing is also suitable for ranging-based tracking~\cite{raab1979em}, with recent work showing their potential for body-worn human pose tracking during rapid movements~\cite{kaufmann2021em}.

\begin{figure}[t]
    \centering
    \vspace{2mm}
    \includegraphics{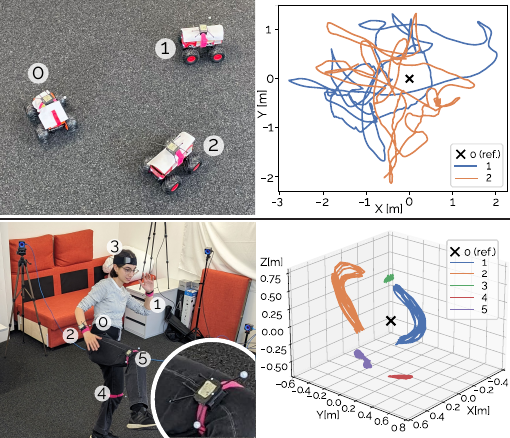}
    \caption{Our tracking system estimates relative positions among a moving group of tracking nodes, demonstrated here for RC cars and pose tracking of a moving person.
    Each tracker has a single-antenna UWB radio, a 6-DoF inertial sensor, and a 3-DoF magnetometer for state estimation.}
    \label{fig:teaser}
\end{figure}

For ranging-based tracking systems, ultra-wideband (UWB) radios are especially promising due to their high bandwidth, time resolution, and compactness~\cite{drone_landing,uloc}. 
UWB ranging can build on signal characteristics such as received signal strength (RSS~\cite{rss}), time of arrival (TOA~\cite{yu2004uwb}), or angle of arrival (AOA~\cite{hol2009tightly}).
RSS offers comparably lower ranging accuracy, whereas AOA ranging requires multiple antennas.
Many approaches have thus been designed based on TOA, often leveraging time difference of arrival, which does not require clock synchronization across receivers~\cite{LedergerberHD15}.
Independent of the ranging scheme, UWB signals are prone to multipath effects and non-line-of-sight distortions caused by the environment, that harm the accuracy of UWB-only positioning systems.
These errors are typically addressed using sensor fusion algorithms, such as Kalman~\cite{welch1995kalman} or particle filters~\cite{ristic2003particlefilters} to incorporate data from other sensors~\cite{idsc_ekf, Xu2021omni, everything_ekf}.
Alternatively, Multidimensional scaling (MDS)~\cite{mds-map} has been used for positioning systems for multiple UWB-only trackers, as standalone implementations~\cite{macagnano2008tracking, Wu2011Robust}, or followed by additional filtering steps~\cite{strohmeier2018pose}.

Crucially, most UWB ranging-based systems typically rely on \emph{multiple antennas} for trilateration~\cite{ait,corrales2008hybrid} or on a stationary tracker~\cite{cao2020accurate,mohammadmoradi2019reflected} as anchor to localize moving 3D trackers (e.g., for robot localization~\cite{drone_landing,drone_fleet}). 
Our recent research explored infrastructure-free tracking with single-antenna UWB wearables for human pose estimation~\cite{uip}.
While we showed that a specifically trained graph neural network can estimate accurate body poses, the technique lacks the flexibility to accommodate \emph{varying} numbers of trackers or generalize to diverse tracking applications.


In this paper, we propose a novel system for collaborative tracking that integrates a \emph{single-antenna} ultra-wideband radio with a 9-DoF magnetic and inertial measurement unit.
Our system affords continuous operation on \emph{constantly moving} tracking nodes, enables each node to opportunistically join and leave the tracked group, and does not depend on any external infrastructure or stationary tracking anchors.

Our tracking method fuses the magnetic and inertial sensor with the pair-wise distances estimated from UWB through  an Extended Kalman Filter (EKF) for each pair of trackers in a constellation. 
Our method then leverages the estimates across sensor pairs to derive relative positions via multidimensional scaling.
A key novelty of our method is the integration of a feedback loop that stabilizes individual EKF instances with the output from MDS to achieve spatial consistency.

\begin{figure}[t]
    \centering
    \vspace{2mm}
    \includegraphics{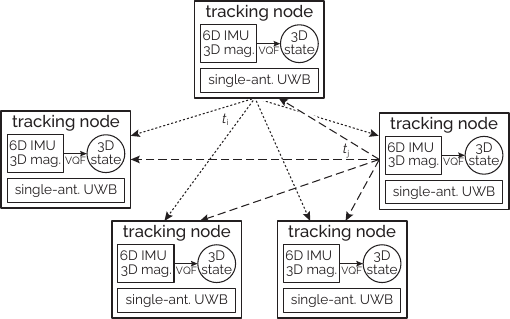}
    \caption{Overview of our tracking system.
    Each tracking node estimates its own 3D orientation state through an embedded VQF filter.
    Through UWB broadcasts, here at $t_i$ and $t_j$, all nodes estimate pairwise distances, which serve as observations in the update step of our Extended Kalman Filter to estimate relative positions.}
    \label{fig:tracking_concept}
\end{figure}

We evaluated our system in two relative tracking applications.
In the first, our system tracked three moving RC cars in environments with and without physical obstacles that occluded the line of sight between cars.
In the second, we attached six trackers to a moving person's body for motion tracking in 3D.
Our system estimated relative tracker positions with an average error of 10.2\,cm (2D relative layout of RC cars on the floor) and 21.7\,cm (3D relative layout for body poses), all while tracked nodes were continuously moving.
Update rates depended on the number of participating trackers, from 81.8\,Hz (three nodes) to 28.6\,Hz (six nodes).
In summary, our contributions are:
\begin{itemize}
    \item a novel tracking system that integrates single-antenna UWB radios with 9-DoF IMUs, enabling compact and versatile tracking nodes,
    \item a multi-stage filtering pipeline that fuses UWB ranging data with inertial measurements through pairwise EKFs, complemented by a novel MDS correction step for enhanced spatial consistency,
    \item a custom ranging protocol that enables opportunistic joining and leaving of tracking nodes, and 
    \item an evaluation in 2D and 3D tracking scenarios with LOS and NLOS conditions. 
\end{itemize}

\section{Method}
\label{sec:implementation}

\subsection{System Overview}
\begin{figure*}[t]
    \vspace{2mm}
    \centering
    \includegraphics{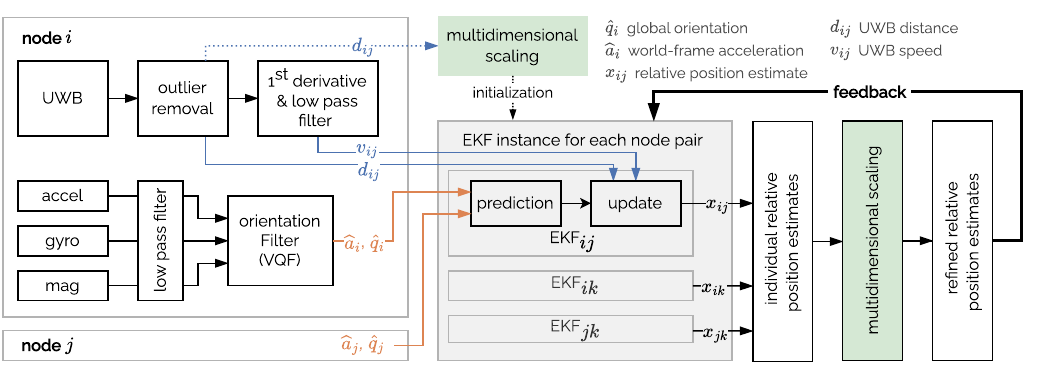}
    \caption{Our pipeline includes individual Extended Kalman Filters for each tracking node's orientation as well as EKFs for each pair of nodes for relative positions.
    A key part of our pipeline is the feedback loop that stabilizes individual EKF instances with the output from MDS to achieve spatial consistency.}
    \label{fig:filtering_pipeline}
\end{figure*}

Our proposed tracking system comprises a set of at least three nodes, all of which can be moving or stationary (Fig.~\ref{fig:tracking_concept}).
Tracking nodes can opportunistically join or leave the cohort;
as soon as they are in range, they participate in our relative position tracking.
Our tracking pipeline comprises three steps to obtain relative positions described in Fig.~\ref{fig:filtering_pipeline}

\subsection{Local state estimation}

In the first step, each node runs an onboard VQF filter~\cite{laidig2023vqf} from IMU and magnetometer readings to estimate its absolute orientation quaternion~$\boldsymbol{\hat{q}}$. With this estimate, we also compute the gravity-compensated world-frame acceleration $\boldsymbol{\hat{a}}$.
As such, each node outputs measurements modeled as follows:
\begin{align}
\boldsymbol{\hat{q}} & = \boldsymbol{q_w} + \boldsymbol{n}_{q} \;,\;  \boldsymbol{n}_{q} \sim \mathcal{N}(0,\boldsymbol{Q}_q)
\label{eq:quat_model}
\\
\boldsymbol{\hat{a}} & = \boldsymbol{a_w} + \boldsymbol{n}_{a} \;,\; \boldsymbol{n}_{a} \sim \mathcal{N}(0,\boldsymbol{Q}_a)
\label{eq:accel_model}
\end{align}
where $\boldsymbol{a_w} = \boldsymbol{q}_w^{-1}(\boldsymbol{a} - \boldsymbol{g})\boldsymbol{q}_w$ and $\boldsymbol{q_w}$ are the ideal world frame acceleration and orientation, $\boldsymbol{g}$ is gravity and $\boldsymbol{Q}_a$ and $\boldsymbol{Q}_q$ are the diagonal covariance matrices used to model additive zero-mean Gaussian noise. As the onboard VQF filter does not explicitly output orientation covariances, both $\boldsymbol{Q}_q$ and $\boldsymbol{Q}_a$ are assumed constant and determined empirically.

\subsection{Pairwise distance measurements}
\begin{figure}[t]
    \centering
    \includegraphics[width=\widthnarrow]{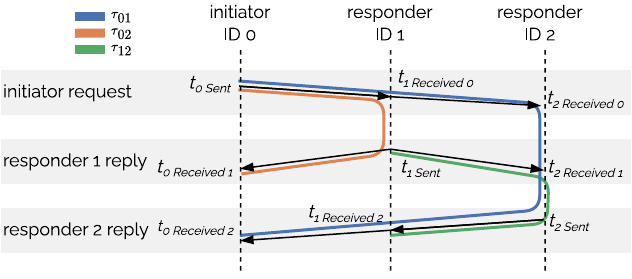}
    \caption{Example of a ranging transaction with two responders. The timestamps required to resolve time-of-flight are included in the UWB message payload and thus broadcasted to all participants in the constellation.
    }
    \label{fig:ranging_protocol}
\end{figure}

We propose a broadcasting-based UWB ranging protocol to estimate pairwise distances within the constellation of trackers.
Range estimates are based on single-sided two-way ranging, in which a request from an initiator, and a reply from a responder are required to estimate the time-of-flight for a device pair. As all UWB messages are broadcast between nodes, we can leverage passive listening to resolve range estimates between responders (Fig.~\ref{fig:ranging_protocol}).
The ranging message payload contains the timestamps, used to resolve time-of-flight $\tau_{ij}$ between two devices. 
\begin{align}
\tau_{ij}  = \frac{1}{2}(t_{i_{Received}} - t_{i_{Sent}} - (t_{j_{Sent}} - t_{j_{Received}}))
\label{eq:uwb_dist}
\end{align}
With this scheme, the overall ideal ranging frequency, which we define as the rate at which all pairwise distances are obtained, is given by:
\begin{align}
f_{SSTWR} = \frac{1}{t_{msg}n}
\label{eq:uwb_cal}
\end{align}
where $n$ is the number of active nodes $t_{msg}$ is the transmission time of a single message.

Our protocol can adapt the ranging frequency to the number of devices on the fly, using a dynamically updated routing table and priority assignment mechanisms to manage ranging roles and delays across nodes. 

Our system assigns each node a unique ID and starts each tracker in initiator mode, periodically broadcasting ranging requests and waiting for replies from responders.
When a node receives any message signed with a smaller ID than its own, it switches to responder mode.
All nodes maintain a routing table with the number and sorted IDs of all active participants, which are then used to adapt ranging delays. 
A watchdog timer resets a device if no UWB messages are received after a timeout, making it possible to recover from an initiator dropout.

\subsection{Sensor calibration}
Each node's IMU is calibrated by estimating initial accelerometer and gyroscope offsets at rest. The magnetometer hard and soft-iron calibration parameters are pre-computed and saved locally.

UWB calibration is required to correct for hardware-specific variables, notably antenna delay and variance of transmission power that significantly affect ranging accuracy. We calibrate all nodes at once, using a RANSAC regression to find an affine mapping of the raw ranges to the ground truth captured with a marker-based motion capture system, maximizing the distance ranges recorded. From this process, we can model a range estimate $\hat{d}$ with Eq.~\ref{eq:uwb_model}. The parameters determined through this process remain valid given similar device operating conditions~\cite{apnote_11}. 
\begin{align}
\hat{d} & = a d + b + \boldsymbol{n_{d}} \;,\;  \boldsymbol{n}_{d} \sim \mathcal{N}(0,\sigma_d)
\label{eq:uwb_model}
\end{align}
where ${d}$ is the real distance, $a$ and $b$ are the affine mapping coefficients, and measurement variance $\sigma_d$.
This model is valid under the assumption of direct line-of-sight (LOS) between devices and ranges under 8 meters~\cite{apnote_14}, in which our evaluation scenarios fall. 

\subsection{Distributed relative position estimation}
The second stage of our tracking pipeline fuses the outputs of the local state estimation stage with the pairwise UWB ranging measurements from tracking nodes to estimate relative positions. We implement an independent Extended Kalman Filter~\cite{welch1995kalman} instance for each pair of devices in the constellation. 

Each filter instance tracks the relative 3D position $\boldsymbol{x}_{ij}$, linear velocity $\boldsymbol{v}_{ij}$ and orientation quaternion $\boldsymbol{q}_{ij}$ between two devices, resulting in the state vector in Eq.~\ref{eq:state_space} 
\begin{align}
\mathbf{x} =
\begin{bmatrix}
x_{ij} \\
v_{ij} \\
q_{ij} \\
\end{bmatrix} = 
\begin{bmatrix}
x_{j} - x_{i} \\
v_{j} - v_{i}\\
q_{i}^{-1}q_{j}\\
\end{bmatrix}
\label{eq:state_space}
\end{align}

In the prediction step, our filter relies on dead-reckoning using the world-frame acceleration $\boldsymbol{\hat{a}}$ and global orientation $\boldsymbol{\hat{q}}$ as control inputs (Eq.~\ref{eq:control_inputs}).


\begin{align}
\mathbf{u} & =
\begin{bmatrix}
\boldsymbol{\hat{a}}_{i} & \boldsymbol{\hat{a}}_{j} & \boldsymbol{\hat{q}}_{i} & \boldsymbol{\hat{q}}_{j}
\end{bmatrix}^\top
\label{eq:control_inputs}
\end{align}

The predicted state $\mathbf{\hat{x}}_k$ at discrete time step $k$ is then expressed as 
\begin{align}
\mathbf{\hat{x}}_k&=f(\mathbf{x}_{k-1}, \mathbf u_k)\\
&=
\begin{bmatrix}
\boldsymbol{x}_{ij_{k-1}} + \Delta T\boldsymbol{v}_{ij_{k-1}} + \frac{\Delta T^{2}}{2} (\boldsymbol{\hat{a}}_{j_k} - \boldsymbol{\hat{a}}_{i_k}) \\
\boldsymbol{v}_{ij_{k-1}} + \Delta T ( \boldsymbol{\hat{a}}_{j_k} - \boldsymbol{\hat{a}}_{k_t})  \\
\boldsymbol{\hat{q}_{i_{k}}}^{-1}\boldsymbol{\hat{q}_{j_{k}}}\\
\end{bmatrix} 
\label{eq:prediction_step}
\end{align}
where $\Delta T$ is the difference between consecutive time steps.

The noise in the prediction step is mainly introduced by the control input. From the sensor models defined in Eq.~\ref{eq:quat_model}--\ref{eq:accel_model}, we define the input's spectral noise covariance matrix $\boldsymbol{\Sigma}_u$ by combining the individual sensor's noise covariance matrices $\boldsymbol{Q}_i$. The prediction step noise covariance $\boldsymbol{Q}_k$ at timestep $k$ can then be expressed as:
\begin{align}
\boldsymbol{Q}_k = \boldsymbol{W}_k\Sigma_u\boldsymbol{W}_k^T
&  &
\boldsymbol{W}_k = \frac{\partial f(\mathbf{x}_{k-1}, \mathbf u_k)}{\partial\mathbf u}
\end{align}

We introduce UWB range measurements in the EKF's correction step, as indirect observations of relative position and velocity. Outliers are removed by comparing an incoming range measurement to a running average of past measurements.
The acceptable variation is dynamically determined using a threshold calculated from the last relative velocity estimate, allowing for adaptive filtering based on current movement speeds.
We then compute the first derivative of the UWB ranges and pass it through a low-pass filter to obtain a speed measurement.
We define the filter's measurement model $\mathbf{h}(\mathbf{x})$ in Eq.\,\ref{eq:h}, and linearize it as Eq.\,\ref{eq:H}.
\begin{align}
\mathbf{h}_k(\mathbf{x}) & =   
\begin{bmatrix}
d \\
v \\
\end{bmatrix} = 
\begin{bmatrix}
\| \boldsymbol{x}_{ij} \|_2 \\
\| \boldsymbol{v}_{ij} \|_2 \\
\end{bmatrix}
\label{eq:h}
\\
H(\mathbf{x}) & = \frac{\partial h(\mathbf{x})}{\partial\mathbf x} =
\begin{bmatrix}
\frac{\boldsymbol{x}_{ij}}{x}&\boldsymbol {0}_3 \\
\boldsymbol{0}_3 & \frac{\boldsymbol{v}_{ij}}{v} \\
\end{bmatrix}_{2\times6}
\label{eq:H}
\end{align}
The measurement covariance matrix $R$ used in the correction step is the diagonal matrix of the variances calculated from empirical data. 


\subsection{Relative position refinement}
 
Individually, each EKF instance is highly sensitive to initialization. As distance measurements introduced in the correction stage are not a direct observation of the state, we expect the relative position estimates to drift over time. However, with a constellation of three or more devices, we can leverage constellation-wide relative position information to constrain drift. As such, we build on multidimensional scaling (MDS) to initialize new EKF instances and refine their individual relative position estimates $\boldsymbol{x}_{ij}$.  

Given $n$ trackers, we first construct a distance matrix $D$, which is symmetric and where $D_{ij}$ is the distance from tracker $i$ to tracker $j$, 
$\boldsymbol{\hat{D}}_{ij} = \| \boldsymbol{x}_{ij} \|_2$.
Due to its symmetry, the matrix has $\tfrac{n}{2}(n-1)$ unique entries.

Next, we aim to represent relative positions such that the pairwise distances between these points match the given distance matrix as closely as possible.
We convert the distance matrix $D$ into a kernel matrix
\begin{align}
    K & = -\tfrac{1}{2} J D^2 J
\end{align}
where the centering matrix $J = I - \tfrac{1}{n}e e^T$ and $e$ is a column vector of ones.

We then obtain the eigenvalues and eigenvectors of $K$:
Let the eigenvalues of $K$ be $\lambda_1 \geq \lambda_2 \geq \ldots \geq \lambda_n$.
Let their corresponding eigenvectors be $e_1, e_2, \ldots, e_n$.
We now solve for the 3D positions,
\begin{align}
    \mathbf{E}_3 & = \begin{bmatrix}
        | & | & | \\
        e_1 & e_2 & e_3 \\
        | & | & | \\
    \end{bmatrix} \\
    \Lambda_3 & = \begin{bmatrix}
        \sqrt{\lambda_1} & 0 & 0 \\
        0 & \sqrt{\lambda_2} & 0 \\
        0 & 0 & \sqrt{\lambda_3} \\
    \end{bmatrix} \\
    X &= \mathbf{E}_3 \Lambda_3^{1/2}
\end{align}
where $\mathbf{E}_3$ is the matrix composed of the eigenvectors corresponding to the three largest eigenvalues, and $\Lambda_3$ is the diagonal matrix with the square roots of these eigenvalues.

We thus obtain our estimated positions matrix $X$, where each row corresponds to the refined estimated position of a tracker in the 3D space.
We use these refined estimates to correct the $\boldsymbol{x}_{ij}$ relative position state for the next iteration of individual EKF instances. 

This architecture also supports the dynamic leaving and joining of tracking nodes. Upon detecting a new tracker, we apply MDS on the first $k$ measurements to determine the initial relative position $\boldsymbol{x}_0$. Similarly, the first $k$ orientation estimates are used to calculate $\boldsymbol{q}_0$, while we set the initial relative velocity $\boldsymbol{v}_0$ and assume an initial identity covariance matrix $\mathbf{P}_0$. As such, we can initialize an EKF instance for each newly formed tracker pair. Conversely, when a tracker drops out of the constellation, we destroy its associated EKF instances. By computing a new distance matrix $D$ at every MDS refinement step, we ensure the entire filtering pipeline adapts to the tracking constellation size. 

\begin{figure}[t]
    \centering
    \vspace{2mm}
    \includegraphics[width=\linewidth]{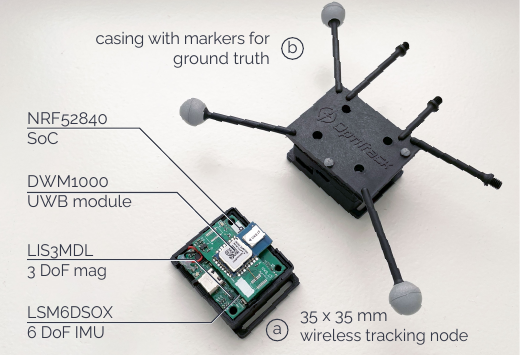}
    \caption{Each embedded tracking node features a base micro-controller PCB with inertial and magnetics sensing, with an extension for UWB ranging.}
    \label{fig:devices}
\end{figure}

\subsection{Hardware implementation of tracking nodes}

We developed custom trackers integrated into a 35$\times$35\,mm package shown in Fig.~\ref{fig:devices}.

Each tracker features an nRF52840 microcontroller, an LSM6DSOX 6-DoF inertial sensor, an LIS3MDL 3-axis magnetometer, and a DWM1000 UWB module.
The IMU and magnetometer are configured to provide readings at the full 16-bit precision (representing $\pm8$\,g, 2000\,mdps, and $\pm8$\,gauss, respectively) at an update rate of 100\,Hz.
We adjusted the parameters of the DWM1000 module to prioritize robustness in multipath environments and line-of-sight occlusions over long-range performance.
Our settings further include a low data rate (110\,kbps), a long preamble sequence (1024\,bytes), and a pulse repetition rate of 16\,MHz (manufacturer recommendations~\cite{apnote_11}).

The ranging protocol and resolution of pairwise distances, as well as the VQF orientation filter, are handled by the microcontroller. The generated data is streamed to a Linux host computer where the rest of the filtering pipeline is implemented using the ROS2 middleware, dynamically spawning and terminating EKF instances for every pair of active devices in the cohort. 
The entire state estimation pipeline runs on a 100\,ms delayed time horizon, accumulated from the various upstream filtering and fusion steps.

\section{Experimental Results}

\subsection{Experiments}

We tested the responsiveness of our system, i.e., how the overall ranging frequency changes with the number of devices in range.
For this, we programmed trackers to turn on or off during an active ranging interval to determine the time needed for the optimal configuration as well as the recovery time in response to a responder or initiator joining or dropping the cohort. 

We then evaluated the performance of our method comparing the relative position estimates with those reported by an optical tracking system. 

\begin{figure}[b]
    \centering
    \includegraphics[width=\linewidth]{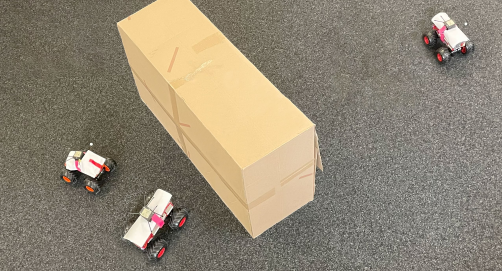}
    \caption{The RC cars drove around an obstacle to simulate intermittent obstructions that cause an NLOS condition.}
    \label{fig:rccar_obstacle}
\end{figure}
Two application scenarios framed the evaluation of our ranging trackers (Fig.~\ref{fig:teaser}).
In the first task, our system tracked remote-controlled cars in 2D while they constantly moved through the space.
We affixed one tracker each to three RC cars and recorded data for several minutes---with and without an obstacle obstructing the line of sight between two cars.  

In the second scenario, we evaluated the performance during a 3D human body motion tracking task.
We attached six trackers to a participant, one each on the right upper arm, right wrist, right hip, left wrist, right knee, and left knee.
The participant performed various free-form motions, including walking, jumping jacks, squats, and so on.

Both experiments were conducted indoors in an area measuring about 3\,m $\times$ 5\,m monitored by 20 Optitrack cameras to record ground truth positions and orientations with sub-millimeter accuracy after calibration. 

\subsection{Results}

\subsubsection{Protocol performance}
\begin{figure}[t]
    \centering
    \vspace{2mm}
    \includegraphics[width=\widthnarrow]{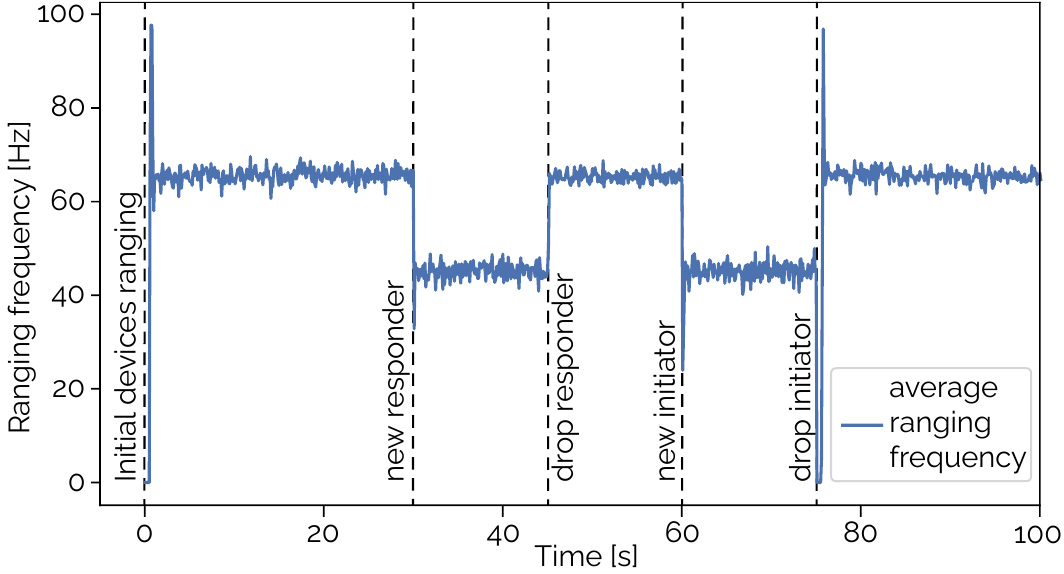}
    \caption{Averaged overall ranging frequency in response to ranging cohort changes. The starting configuration has 4~devices, with a new responder joining and dropping, followed by a new node taking over the initiator role before dropping out as well. 
    }
    \label{fig:ranging_freq}
\end{figure}

The results of our protocol responsiveness test are plotted in Fig.~\ref{fig:ranging_freq}.
Responder events have a negligible impact on the ranging protocol with a response delay of 12.3\,ms on average for the protocol to stabilize.
The initiator dropping out is equivalent to a system reset that requires 733.4\,ms on average to recover from.

\begin{figure*}[t]
    \centering
    %
    %
    \includegraphics[width=\linewidth]{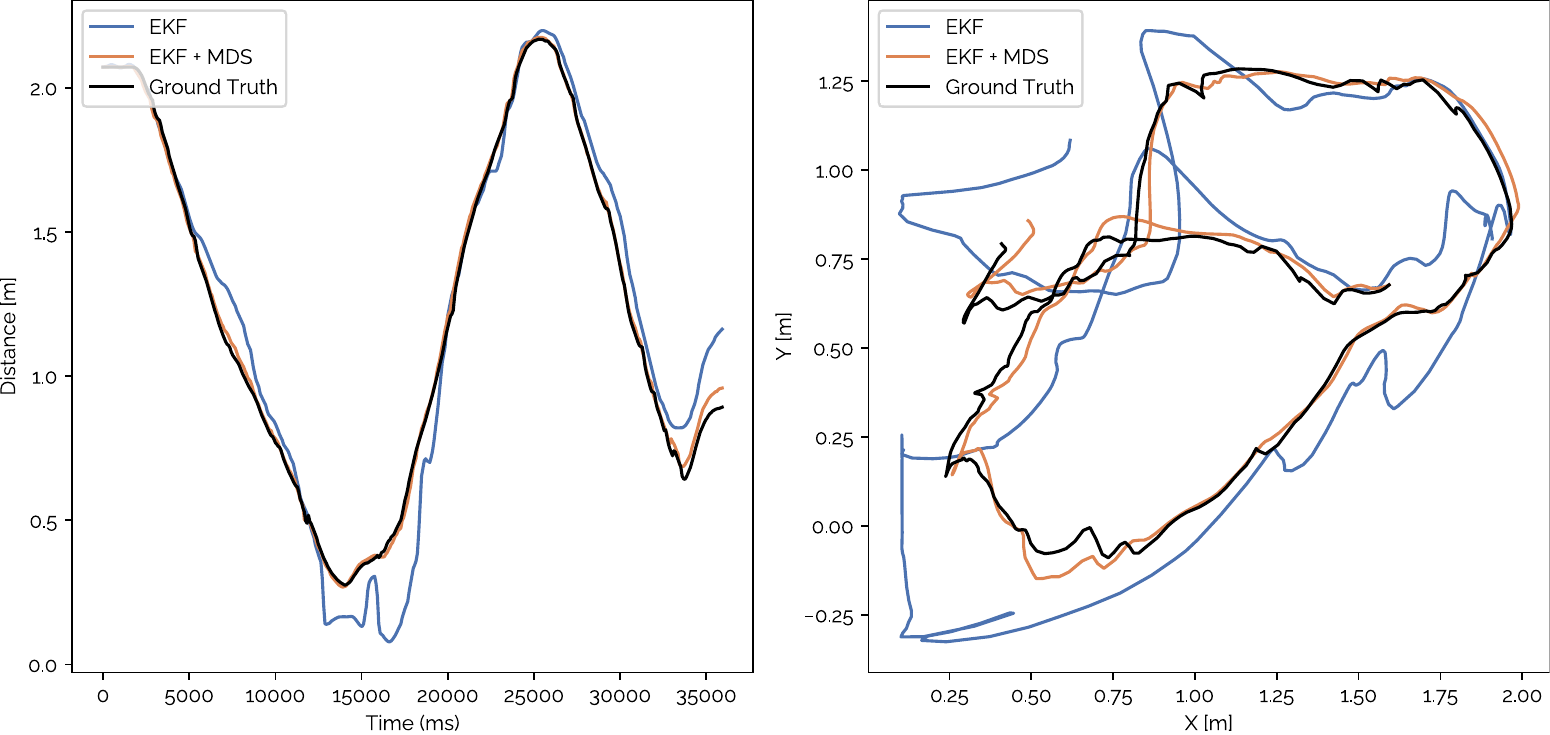}
    \caption{Sample of relative position estimates $\boldsymbol{x}_{ij}$ and corresponding distances $\| \boldsymbol{x}_{ij} \|_2$ between Cars 0 and 2 in our NLOS evaluation with RC cars.}
    \label{fig:mainfig}
\end{figure*}

\subsubsection{Position estimation}
\begin{table}
    \vspace{1mm}
    \centering
    \begin{tabular}{lrrr}
    \hline &&& \\[-2ex]
      & RC cars (LOS) & RC cars (NLOS) & Human motion \\
      & 3 trackers, 2D & 3 trackers, 2D & 6 trackers, 3D \\
    \hline &&& \\[-2ex]
    EKF only  & 39.6\,cm & 42.8\,cm & 66.7\,cm\\
    EKF + MDS & 9.3\,cm & 10.2\,cm & 21.7\,cm\\
   \hline
    \end{tabular}
    \vspace{1mm}
    \caption{Relative position error (RMSE)}
    \label{table:results}
\end{table}


The tracking experiment results, shown in Table~\ref{table:results}, indicate our system performs better in the RC car tasks. While this application benefits from being constrained to 2D, other factors also contribute to this difference in performance. First, with only 3 participating trackers, the RC car experiments have the advantage of a higher ranging frequency (81.83\,Hz) compared to the obtained with 6 body trackers (28.6\,Hz). A higher UWB ranging frequency implies more frequent EKF correction steps and less drift accumulating from IMU dead-reckoning. Second, the two experiments differ in the nature of NLOS conditions they face. In the RC car experiment, NLOS situations are intermittent and only affect the raw UWB ranging inputs when the line of sight between 2 moving cars is temporarily occluded by the obstacle and the filtering pipeline relies on UWB range measurements with variance $\sigma_d = 11.6$\,cm. Meanwhile, some device pairs in the human tracking case are consistently occluded by the body, such as the head-knee pairs, for which UWB ranging noise reaches $\sigma_d = 27.5$\,cm. 

Also highlighted in our results is the contribution of the MDS refinement step to constrain errors in relative position estimation. Fig.~\ref{fig:mainfig} shows that while distance estimates $\| \boldsymbol{x}_{ij} \|_2$ do not diverge with the EKF-only estimation, the corresponding relative position estimate error compounds over time without MDS correction. 

\section{CONCLUSION}

We have presented a relative tracking system for opportunistic constellations of tracking nodes, each of which integrates a single-antenna ultra-wideband radio with a 9-DoF magnetic and inertial measurement unit.
Our system can track a distributed constellation of constantly moving nodes that nodes can spontaneously join or leave.
Our system implements a multistage filtering pipeline to estimate all relative positions within the constellation of trackers.
Our pipeline first estimates relative positions between all sensor pairs individually, then integrates multidimensional scaling-based coordinates to refine estimates using the collective information across the constellation.
We evaluate our system with 2D and 3D tracking scenarios, achieving tracking accuracies of 0.102\,m and 0.217\,m, respectively, without introducing the constraint of a static anchor.







\balance{}
\bibliographystyle{z-references/IEEEtran}
\bibliography{z-references/swarm,z-references/rayan-thesis,z-references/new_refs}

\end{document}